%% file: main.tex
\documentclass[10pt,journal,compsoc]{IEEEtran}

\usepackage{graphicx}
\usepackage{booktabs}
\usepackage{subcaption}
\usepackage{comment}
\usepackage{enumitem} %
\usepackage{amsmath}
\usepackage{amssymb}
\usepackage{color}
\usepackage{xcolor}
\usepackage{cite}
\usepackage[hide links,bookmarks=false]{hyperref}
\iftrue
\makeatletter
\renewcommand{\paragraph}{%
  \@startsection{paragraph}{4}%
  {\z@}{1.75ex \@plus 1ex \@minus .2ex}{-1em}%
  {\normalfont\normalsize\bfseries}%
}
\makeatother
\fi

  \newcommand{\etal}{\textit{et al. }}
\newcommand{\example}[1]{\textit{#1}}
\newcommand{\jacob}[1]{\textcolor{blue}{[#1 -J]}}

\renewcommand{\emph}[1]{\textbf{#1}}
\newcommand{\varname}[1]{\textsc{#1}}

\begin{document}

\title{\#anorexia, \#anarexia, \#anarexyia: Characterizing Online Community Practices with Orthographic Variation} 

%\numberofauthors{4}
\author{
\IEEEauthorblockN{Ian Stewart, Stevie Chancellor, Munmun De Choudhury, and Jacob Eisenstein\\}
\IEEEauthorblockA{Georgia Institute of Technology\\
Atlanta, GA\\
\texttt{\{istewart6,schancellor3,munmun.choudhury,jacob.eisenstein\}@gatech.edu}} 
%\vspace{-25pt}
%801 Atlantic Drive NW\\
%Atlanta, GA 30318\\ }
}

\maketitle

\begin{abstract}
Distinctive linguistic practices help communities build solidarity and differentiate themselves from outsiders. 
In an online community, one such practice is variation in \textit{orthography}, which includes spelling, punctuation, and capitalization.
Using a dataset of over two million Instagram posts, we investigate orthographic variation in a community that shares pro-eating disorder (pro-ED) content. 
We find that not only does orthographic variation grow more frequent over time, it also becomes more profound or ``deep,'' with variants becoming increasingly distant from the original: as, for example, \example{\#anarexyia} is more distant than \example{\#anarexia} from the original spelling \example{\#anorexia}.
We find that the these changes are driven by newcomers, who adopt the most extreme linguistic practices as they enter the community. 
Moreover, this behavior correlates with engagement with the community: the newcomers that adopt deeper variant orthography tend to remain active for longer in the community, and posts with deeper variation receive more positive feedback in the form of ``likes.'' 
Previous work has linked community membership change with language change, and our work casts this connection in a new light, with newcomers driving an evolving practice rather than adapting to it. 
We also demonstrate the utility of orthographic variation as a new lens to study sociolinguistic change in online communities, particularly when the change results from an exogenous force such as a content ban.
\end{abstract}

\IEEEpeerreviewmaketitle

\input{intro}

\input{related}
\input{data}
\input{methods}
\input{results}
\input{discussion}
\input{conclusion}
\input{acknowledgments}

\bibliographystyle{IEEEtran}
\bibliography{main}

\end{document}

%% file: intro.tex
\section{Introduction}
Online communities are defined by their membership and the shared practices of their members. 
A member of a community with strictly civil practices, like thanking someone for answering a question, would likely have trouble adapting to the language of a community like 4chan~\cite{bernstein2011}. 
The adoption of such practices can differentiate new members from regular community members, as new members must learn the community's practices in order to be considered a regular community participant~\cite{bryant2005,lave1991}.
Among community practices, language plays a particularly important role as a signal of shared identity~\cite{labov2001}.
In the online setting, nonstandard \textbf{orthography} such as ``leet speek'' can differentiate community newbies or ``noobs'' from accepted members~\cite{androutsopoulos2011}.
As important as language practices are, they are subject to constant change as a result of exogenous and endogenous events~\cite{danescu2013,kooti2012emergence}. 
Who in a community drives these changes? 
If changing practices are not adopted by all community members, then what characterizes the members who accept and advance these changes?

%% This paragraph still needs work, come back to it
The social meaning of language change in online communities can be better understood by linking language change to community membership dynamics, i.e., the progression of individual community members from new to regular member. 
For example, studies have shown that the adoption of slang words and jargon
follows predictable temporal patterns, both at the community level and over the lifespan of individual community members~\cite{danescu2013,eisenstein2014}. 
This lifecycle pattern mirrors the generational aspect of language change by which children acquire a dialect from their parents and peers, and then retain the dialect into adulthood (the ``adult language stability assumption'')~\cite{labov2001}.
However, language change may also result from exogenous shocks, such as a content ban in an online community~\cite{chancellor2016variation,hiruncharoenvate2015}.
In 2012, Instagram banned hashtags that promoted eating disorder behaviors, or pro-ED content, such as \example{\#thinspo}~\cite{hasan2012}. 
In response, members of the pro-ED community adopted orthographic variations of hashtags to circumvent the ban. 
Over time, these hashtags grew more popular and more complex, becoming increasingly distant from the original spellings.

This paper outlines a novel approach to measuring change in community practices via orthographic variation, exploring the following three research questions:
\begin{itemize}
\item \emph{RQ1}: Who uses orthographic variants? 
\item \emph{RQ2}: Is depth of variation affected by membership attributes (i.e. age and lifespan)?
\item \emph{RQ3}: Does orthographic variation affect social reception (via likes and comments) of pro-ED content?
\end{itemize}

We first address the correlation of orthographic variation to the behavior of pro-ED community members and then the social reception of such variation. 
In RQs 1 and 2, we focus on two variables that define community membership: \emph{age} in the community and \emph{lifespan}. 
Prior work has highlighted the role of member age as a factor in the adoption of practices: newcomers can drive adoption of new words within a community but may become more resistant to change as they spend more time in the community~\cite{danescu2013}. 
Furthermore, member lifespan, or total duration of time spent in the community, can impact adoption of community practices~\cite{ren2011}. 
RQ3 addresses the social relevance of orthographic variation, which can help explain its adoption within the community.

To address these questions, we analyze over two million Instagram posts and nearly 700  orthographic variants of pro-ED hashtags on Instagram. 
We find that in this community, orthographic variation is driven primarily by newcomers, especially those who will become long-term participants: these individuals are more likely to use orthographic variants, particularly deep variants that are far from the original spellings. 
The depth of orthographic variation is also correlated with community engagement: messages containing deeper orthographic variants receive more ``likes.'' To assess the impact of Instagram's content ban, we compare against Twitter data from the same time period, finding that only Instagram experienced a rapid growth in orthographic variation.

%% file: related.tex
\section{Related Work}

Our work draws on research on the adoption of community practices and research in language variation that focuses on orthographic variation.

\subsection{Adoption of community practices}
The process of knowledge transfer and community growth can be viewed within the framework of \emph{communities of practice}. 
A community of practice is a group of people who share a set of problems and who demonstrate their expertise in the area through the development of consistent practices~\cite{dube2006}.
Communities of practice relate to the theory of Legitimate Peripheral Participation (LPP)~\cite{lave1991}, under which newcomers learn community practices from older members to become full participants. 
LPP has been frequently employed in qualitative research on online communities to understand the establishment of practices~\cite{bryant2005,schlager2002}. 

Under LPP, new community members begin at the periphery and become a full participant through the adoption of community practices~\cite{lave1991}. 
Community practices often receive enforcement from members with more authority or experience~\cite{blashki2005}, such as regular and well-connected members~\cite{kooti2012emergence}. 
However, other studies have shown that newcomers are early adopters of ongoing changes; these individuals then become conservative, maintaining the practices that were innovative at the time when they joined the community~\cite{danescu2013}. 
Community practices may also be adopted differently depending on member lifespan: for instance, transient, or short-lifespan, members often invest less in community practices than committed, or long-lifespan, members~\cite{ren2011}. 

In our study of changing language practices, we argue that the Instagram members who have adopted pro-ED hashtags form a community of practice. 
Shared practices include the use of these variants, which make it possible to share pro-ED content in defiance of Instagram's efforts to moderate. 
Prior work has shown that pro-ED content online is produced by a consistent set of users with similar posting practices and a shared set of problems~\cite{chancellor2016variation,cobb2016,manikonda2017}, which further supports our framing of these Instagram users as a community of practice. 

\subsection{Language Variation}
Language variation refers to structured and consistent differences in language use across communities, individuals, or situations. 
Variation can reveal important social distinctions in attitudes and personal identities~\cite{eckert1992,labov2001}. 
Furthermore, language can vary over time, as when a new generation of speakers learns the language of the previous generation and advances a language change in progress (``transmission'')~\cite{labov2007}. 
Sociolinguists have largely focused on variation in spoken language, but written text also exhibits variation, especially in online settings in which traditional language norms may be relaxed~\cite{eisenstein2013,herring2012}.

Our work examines orthographic variation, the deliberate use of alternative spellings and other character-level features~\cite{jaffe2012orthography}. 
This includes phenomena ranging from phonologically-motivated spellings~\cite{eisenstein2015systematic} to purely typographical practices such as ``leet speak''~\cite{blashki2005} and alternative capitalization schemes~\cite{tsur2015}. 
Orthographic variation online has been tied to a variety of social behaviors such as identity expression~\cite{androutsopoulos2011,zelenkauskaite2008}, stylistic creativity~\cite{schnoebelen2012}, and community membership~\cite{back2013,leigh2009}.  
However, there has been little work demonstrating how orthographic variation arises and spreads through online communities. 
Our work therefore breaks new ground in three important ways: (1) by tracking the use of orthographic variation over each user's lifespan in the community, (2) by linking orthographic variation to signals of social reception, and (3) by explicitly differentiating the frequency and depth of variation.

%% file: data.tex
\section{Data}
%% too many section headings for me -J
%\subsection{Data Context}
\begin{figure*}[t!]
\includegraphics[width=\textwidth]{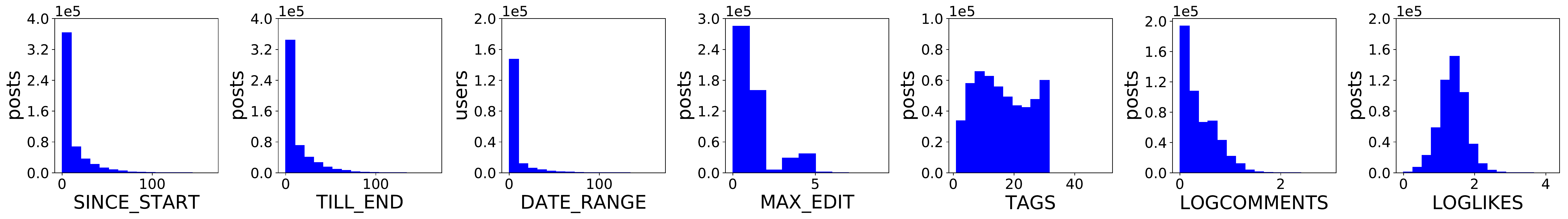}
\caption{Summary histograms for all variables of interest, including relative time (e.g. \varname{date\_range}), linguistic (\varname{max\_edit}) and social variables (\varname{logcomments}).}
\label{fig:summaryStats}
\end{figure*}

We employ a dataset with over two million Instagram posts, gathered from a set of ``pro-ED'' hashtags, which promote disordered eating and exercise behaviors~\cite{chancellor2016variation}. 
This dataset includes manual annotations of the links between hundreds of orthographic variant hashtags and their original spellings.
%Studies in language variation need extensive manual labeling to find social patterns~\cite{labov2001}, and this dataset allows us to quickly label millions of posts. 
Knowing the original spelling for each variant (e.g., that \example{\#anarexyia} is related to \example{\#anorexia}) makes it possible to compute the distance between the variant and source, and thus to quantify the depth of variation.

% combined MEGA-figure!
\begin{comment}
\begin{figure*}[h!] % [htp]
\centering
\sffamily
\small
\begin{minipage}[h]{.3\textwidth}
\includegraphics[width=\textwidth]{figures/variant_post_frequency_over_time.png}
\end{minipage}\qquad
\begin{minipage}[h]{.3\textwidth}
\includegraphics[width=\textwidth]{figures/user_lifespan_distribution.png}
\end{minipage}
\begin{minipage}[h]{.3\textwidth}
    \begin{tabular}{p{1.4cm} p{1.5cm} p{1.5cm} }
    ~                  & All       & Variant-posting \\ \hline
    Members              & 174,601   & 86,540        \\
    Posts              & 2,416,258 & 1,253,406     \\
    $\mu$(Posts) & 13.8      & 14.5          \\ \hline
    \end{tabular}
\end{minipage}
\caption{Summary statistics for variant vs. overall post data. ``Variant-posting members'' includes any member who included at least one variant in one of their posts. \jacob{needs tweaking}}
\label{fig:summaryStats}
\end{figure*}
\end{comment}

\subsection{Data Collection}
%We received a dataset of over two million posts and 689 labeled hashtags with 672 unique orthographic variations of seventeen source tags, originally used by Chancellor~\etal~\cite{chancellor2016variation}.
Details of data collection can be found in the original paper by Chancellor~\etal~\cite{chancellor2016variation}. 
We summarize below only the most relevant points for this research. 

The dataset was acquired in late 2014, using the public Instagram API to search for pro-ED hashtags. Because many hashtags could not be queried directly due to the Instagram bans, Chancellor~\etal{} identified a set of nine non-banned ``seed tags'' related to eating disorders. 
They gathered posts on those seed hashtags for 30 days, and identified the 222 most popular hashtags related to pro-ED behaviors. 
They manually removed hashtags that were ambiguous (e.g. \example{\#fat}) or related to eating disorder recovery (e.g. \example{\#anorexiarecovery}). 
This resulted in a set of 72 hashtags, which they used to gather a large dataset. 
After removing posts with recovery hashtags, the dataset contained 6.5 million posts, dating between January 2011 and November 2014. 

From these 6.5 million posts, Chancellor~\etal manually checked the top 200 most popular hashtags to see how many were banned by Instagram or placed on a ``content advisory''~\cite{hasan2012}. 
They found seventeen source hashtags (e.g. \example{\#thighgap} or \example{\#anorexia}) that underwent some form of Instagram intervention. They then developed a set of regular expressions (e.g. $an*a*$ for \example{\#ana}) to extract semantically similar yet orthographically variant hashtags from the source hashtags. The manual rating yielded 672 unique {\it orthographic variants}, and seventeen source hashtags, totaling 689 hashtags, which we study here. 

In total, the dataset has 2,416,259 posts from January 2011 to November 2014, each of which contains at least one orthographic variant or source hashtag. Of these, 51\% contain at least one variant and no source hashtags.

\begin{table*}[t!]
\centering
\small
% p{7.5cm} p{1.3cm} p{1.3cm}
\begin{tabular}{ l l l l p{3.5cm} } \toprule
Edit distance & Top 3 variants & Source hashtags & Unique variants & \% posts with at least one variant in group \\ \midrule
1 & \example{anarexia, bulimic, eatingdisorders} & 17 & 253 & 41.1\% \\
2 & \example{anarexyia, thinspooo, thynspoo} & 15 & 221 & 2.07\% \\ 
3 & \example{secretsociety123, thinspoooo, thygap} & 15 & 108 & 9.60\% \\ 
4 & \example{secret\_society123, secretsociety\_123, thinspooooo} & 10 & 50 & 10.4\% \\ \bottomrule
% 5 & thygapps & 22 & 10,166 \\ \hline
% 6 & thynspire & 8 & 237 \\ \hline
% 7 & thinspoooooooo & 5 & 23 \\ \hline
% 8 & anorexiaanorexianervosa & 2 & 7 \\ \hline
% 9 & thinspoooooooooo & 1 & 5 \\ \hline
\end{tabular}
\caption{Summary of orthographic variants grouped by edit distance. The edit distance 1 group has the greatest variety of source hashtags and unique variants, while the edit distance 4 group has the lowest variety.
We restrict our study to variant hashtags with edit distance at or below 4, due to data sparsity above edit distance 4.
}
\label{tab:editDistStats}
\end{table*}

\subsection{Feature extraction} 
The following features are extracted from each post and associated Instagram community member:
\begin{itemize}
\item Real time of post, measured in weeks since Instagram instituted a ban on several pro-ED hashtags\footnote{This date is not reported by Instagram but is estimated to be April 1, 2012~\cite{chancellor2016variation}.} (\varname{date}).
\item Number of weeks since the member's initial post in the data, measuring the user's \emph{age} (\varname{since\_start}).
\item Number of weeks until member's final post in the data (\varname{till\_end}).
\item The total duration (in weeks) of a member's activity, measuring the user's \emph{lifespan} (\varname{date\_range}).
\item The appearance (binary) of any variant in the post (\varname{variant}).
\item The appearance (binary) of a variant with a specified edit distance in the post (\varname{edit\_dist\_1}, etc.; see \autoref{subsec:edit_dist} for a description of how edit distance is computed).
\item Maximum orthographic edit distance out of all variants in the post (\varname{max\_edit}); set to 0 when no variants were in post.
\item Total number of all hashtags (variant and non-variant) per post (\varname{tags}).
%\item Maximum popularity out of all variants in post $P$ (\varname{max\_pop}), where ``popularity'' equals the total frequency of the variant within the data at the time of post $P$; log-transformed to adjust for the long tail.
\item Number of comments (\varname{comments}) and likes (\varname{likes}) on a post, counted at time of data collection in 2014; log-transformed to adjust for the distributions' long tails.
\end{itemize}

\begin{figure}[t!]
\includegraphics[width=\columnwidth]{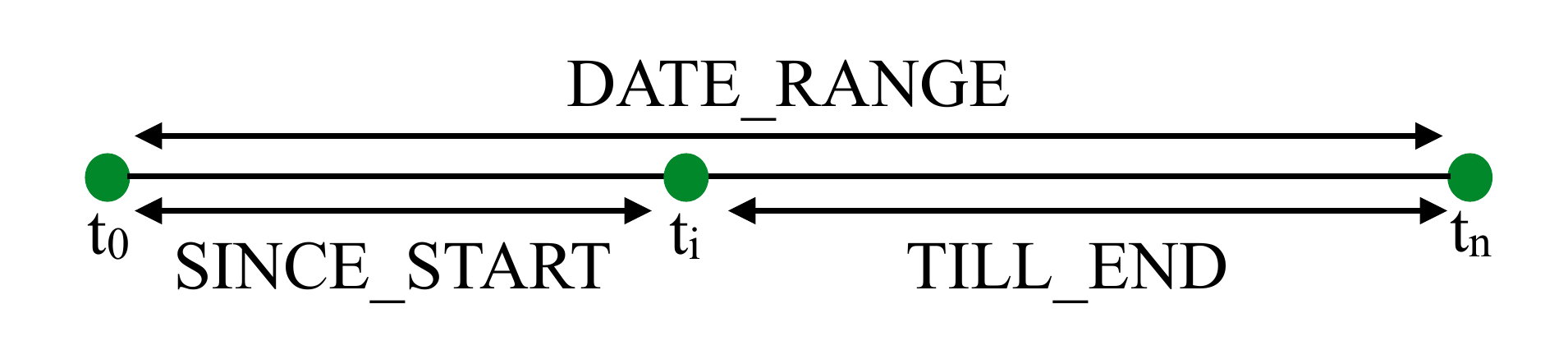}
\caption{Example timeline of member posts at times $t_{0}$ (first), $t_{i}$ and $t_{n}$ (final) that shows age with statistics \varname{since\_start} and \varname{till\_end}, and showing lifespan with \varname{date\_range}.}
\label{fig:time_variable_diagram}
\end{figure}

The distributions of all scalar variables are shown in Figure~\ref{fig:summaryStats}. 
All of the temporal variables have long-tail distributions, indicating that most user lifespans are short.
%% max_pop is not in the figure! -J
% Note the bimodal distribution for \varname{max\_pop}, showing that the overwhelming majority of posts either contain at least one very popular hashtag, or no popular hashtags at all.

%% file: methods.tex
\section{Methods}

We now outline the methods used in our analysis, including operational definitions for key terms, the edit distance metric used to quantify orthographic variation, and the statistical approaches to address our research questions. 

\subsection{Definitions}

For convenience, definitions are provided for the key concepts in our study. 
We refer to individual Instagram users as ``community members'', due to their participation in the pro-ED community, as signaled by the use of pro-ED hashtags.

\begin{itemize}
\item \emph{Age}: for a given post and the associated member, the length of time between the post at time $t_{i}$ and the first pro-ED post created by the member time $t_{0}$. Age is quantified as the variable \varname{since\_start}, which is equal to the number of weeks since the member's first pro-ED post ($\varname{since\_start}=t_{i}-t_{0}$). The variable \varname{till\_end} equals the number of weeks until the member's final pro-ED post at time $t_{n}$ ($\varname{till\_end}=t_{n}-t_{i}$). These statistics are shown in Figure \ref{fig:time_variable_diagram}. We define a \emph{newcomer} as a member who, at time of posting, has spent less than ten weeks in the community, and a \emph{regular} as a member who, at time of posting, has spent at least ten weeks in the community.\footnote{It is possible that some members had additional posts in the time between the launch of Instagram in 2010 and the beginning of our data collection in 2011. This would cause us to underestimate the age of some individuals. However, our dataset spans four years, and the overwhelming majority of members appear to have ages of less than one year.}
\item \emph{Lifespan}: for a given member, the length of time between a member's first and final pro-ED post. Lifespan is quantified as the variable \varname{date\_range}, which is equal to the number of weeks between the member's first and final pro-ED post ($\varname{date\_range}=t_{n}-t_{0}$). This statistic is shown in Figure \ref{fig:time_variable_diagram}. We define a \emph{transient} community member as having a lifespan less than ten weeks in length, and a \emph{committed} member as having a lifespan of at least ten weeks.
\item \emph{Source}: any pro-ED hashtag that was banned in April 2012 and has at least one documented orthographic variant; e.g., \example{\#anorexia}.
\item \emph{Variant}: any orthographically-varied hashtag that can be associated with a source hashtag; e.g., \example{\#anoreksya}.
\item \emph{Depth}: the linguistic distance between a source and its variant: e.g., the variant \example{\#anoreksya} has a depth 3 from its source \example{\#anorexia} (see \autoref{subsec:edit_dist}).
\end{itemize}

We acknowledge that the variables \varname{since\_start}, \varname{till\_end}, and \varname{date\_range} only capture a slice of each community member's behavior, because a pro-ED hashtag member's actual first post on Instagram may be unrelated to pro-ED (and thus unobservable).
%However, knowing that the pro-ED hashtag users on Instagram form a coherent community~\cite{chancellor2016variation}, we can still use the temporal variables defined to draw conclusions about 

\subsection{Measuring Orthographic Variation: Edit Distance}
\label{subsec:edit_dist}

The depth of orthographic variation is quantified by calculating each variant's Levenshtein edit distance from its original form~\cite{levenshtein1966binary}. We count the minimum number of insertion, deletion and substitution operations necessary to convert an source hashtag to its variant form. For example, transforming \example{anorexia} to \example{anoreksya} requires two substitutions ($x \rightarrow k$ and $i \rightarrow y$) and an insertion ($\varnothing \rightarrow s$), thus an edit distance of $1 \times 2 + 1=3$. Although in some cases it is useful to design a customized edit distance cost function~\cite{heeringa2006},
% or derive a cost function automatically~\cite{mccallum2005conditional}, 
%% no point in mentioning this -j
in this study we weight all operations equally for simplicity.\footnote{Preliminary tests with an experimental weighted edit distance showed little difference from the tests with the unweighted edit distance.}

\begin{figure}[t!]
\begin{center}
\includegraphics[width=0.9\columnwidth]{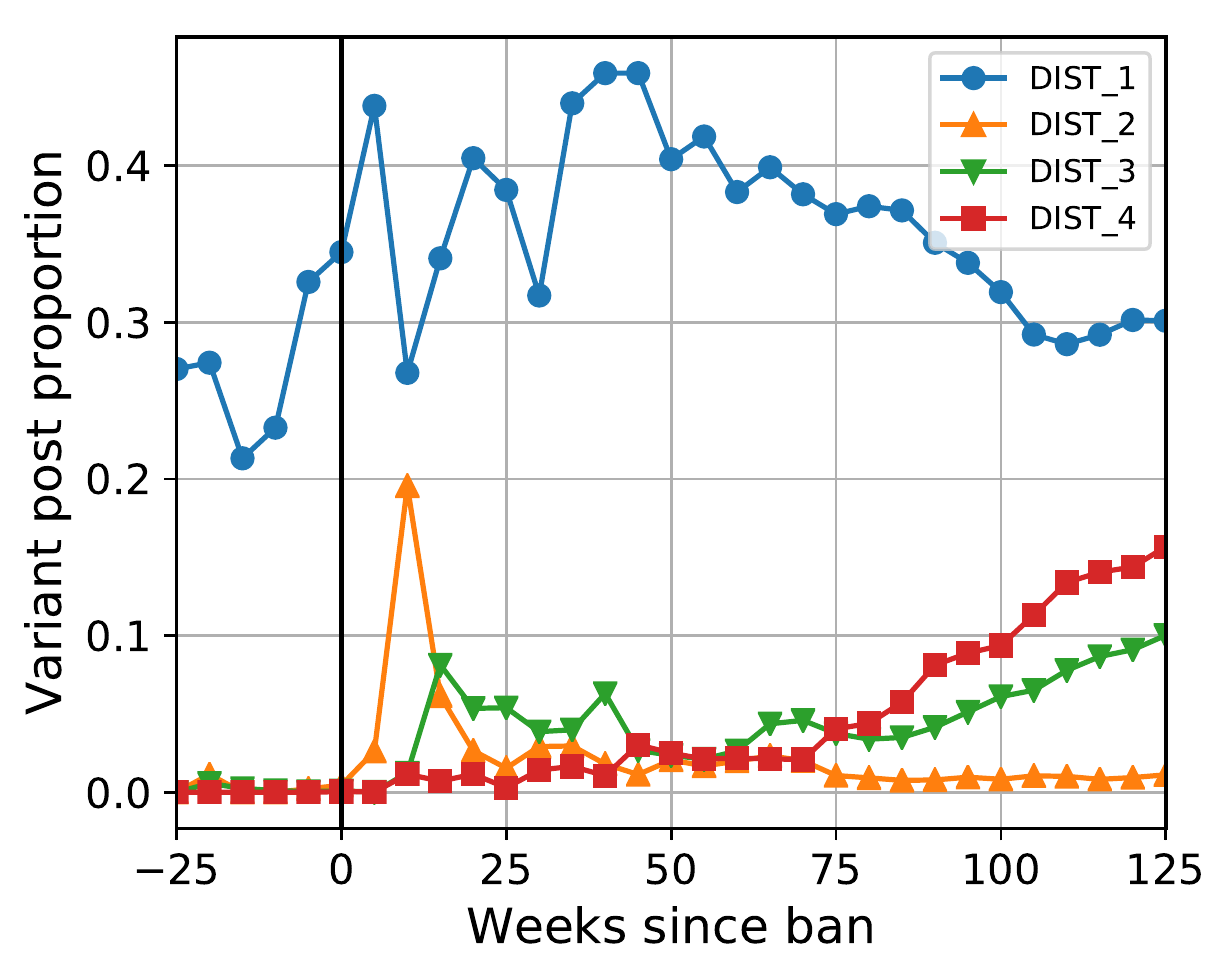}
\caption{Frequency of variants over time, grouped by edit distance: e.g., \varname{dist\_1} tracks the normalized frequency of all posts with at least one variant with edit distance 1, such as \example{\#anorexiaa}. 
}
\label{fig:editDistTime}
\end{center}
\end{figure}

We group orthographic variants by edit distance in Table~\ref{tab:editDistStats} and provide summary statistics for each group, showing the uneven distribution across groups. We also display the frequency of variants grouped by their edit distance in Figure~\ref{fig:editDistTime} and note that the overall frequency of orthographic variants increases over time, particularly for the deeper variants at edit distances 3 and 4.  Our study examines which community members drive this increase in the frequency and depth of variation over time.

\subsection{Statistical Models}

%Our study examines the relationship between orthographic variation and member behavior over time (RQ1). 
%Additionally, we investigate the attributes of community members who drive the adoption of orthographic variation, measured in member age and lifespan (RQ2), and the social reception of such variation, measured in likes and comments on posts (RQ3). 
We use logistic and Poisson regressions as models for their ease of interpretability, since our RQs concern the relative importance of the temporal, social and linguistic variables of interest. 
%Since the dependent variables in RQ1 and RQ2 are binary, logistic regression is an appropriate choice. 
We choose a Poisson regression to address the dependent variables (\varname{likes} and \varname{comments} in RQ3), because they are count variables with high dispersion and non-normal distributions~\cite{gardner1995regression}. 
%For regressions in which the predictors have multiple unit types, we scale the independent variables between 0 and 1 for consistency.\jacob{I thought we standardized?}\footnote{Following ~\cite{chinn2000}, we convert logistic regression coefficients to normal equivalent deviates by dividing by the standard deviation of the standard logistic distribution, $\frac{\pi}{\sqrt{3}} \approx 1.81$. \jacob{The effect sizes are not equal to $\beta/1.81$, how to explain?}} %\jacob{Careful, don't do this when dealing with the coefficients of the linear regression. I'm not sure what to do with Poisson regression coefficients.} 
%For interpretability, the coefficients are shown in log-odds-per-week.
The specific regression models for each RQ are described below:

\textbf{RQ1: Who uses orthographic variants?} 
We use logistic regressions to predict whether a variant spelling appears in a post (dependent variable), using the following membership attributes as independent variables: (1) the post author's relative age (\varname{since\_start} and \varname{till\_end}) and (2) the post author's lifespan (\varname{date\_range}), as well as the absolute time \varname{date} for both variables.

\textbf{RQ2: Is depth of variation affected by membership attributes?} 
Depth of variation is measured as the edit distance of a variant from its original form. 
We consider as a dependent variable the presence of a variant of a specified edit distance from the original tag (e.g., any variant with edit distance 4). 
We again perform a set of logistic regressions, using the same independent variables as in RQ1 to determine the importance of membership attributes. 
%{\color {cyan} We need to concretely define depth of variation here, and what is meant by author properties - MDC}

\textbf{RQ3: Does orthographic variation affect social reception?} 
We use Poisson regressions to predict the number of likes and comments that a post receives (dependent variable), using as independent variables the membership attribute variables of the posting member (\varname{date\_range} and \varname{since\_start}) as well as the post's language content (\varname{tags}, \varname{max\_edit}, \varname{variant}). We include a fixed effect for each member to account for varying popularity among members.

In all regressions, we remove duplicate posts by members who contribute more than one post for each date to avoid overfitting to the most active members. 
For logistic regressions, we randomly subsample the data ($n=200,000$) and include an equal number of positive and negative labeled instances (for class balance). 
We demonstrate the relative goodness of fit of models using the metric \textit{deviance}, which is a measure of the lack of fit to data
%~\cite{hosmer2004applied} 
(lower values are better). 
A model's deviance is calculated by comparing the model with the saturated model,
%---a model with a theoretically perfect fit, 
which we define as the ``null model.'' 
To interpret the relative importance of the variables in the above regression models, we report the non-standardized coefficients of the regression, $p$-values (computed through the Wald test, adjusted for Bonferroni correction), and standardized effect sizes~\cite{chinn2000}. 
All regressions are performed using the Generalized Linear Model code from the \texttt{statsmodels} Python package.\footnote{http://statsmodels.sourceforge.net/stable/glm.html} 

%% file: results.tex
\section{Results}

We address our RQs by analyzing (\autoref{sec:rq1}) the attributes of community members who adopt orthographic variants, (\autoref{sec:rq2}) the correlation between orthographic depth and membership attributes, and (\autoref{sec:rq3}) the correlation between orthographic depth and social reception. 
We also include a comparison with Twitter data to determine the prevalence of orthographic variation across social media platforms (\autoref{sec:twitter_comparison}). In all regressions, the coefficients $\beta$ are expressed in terms of the units of the predictors, e.g. log-odds per week. Effect sizes are computed by standardizing the predictors to zero mean and unit variance, and then dividing the resulting coefficients by $\pi/\sqrt{3}$, the standard deviation of the standard logistic distribution~\cite{chinn2000}.

\subsection{RQ1: Who uses orthographic variants?}
\label{sec:rq1}
The first task is to 
%investigate the relative time variables outlined in RQ1 and 
determine whether a specific subgroup, such as newcomers~\cite{danescu2013}, appears to drive the community-level tendency toward more orthographic variation.
%We use age and lifespan to represent a member's role within the pro-ED community, as a newcomer versus regular and transient versus committed member. 
%The results of the regression help us to investigate whether a specific subgroup, such as newcomers~\cite{danescu2013}, appears to drive the community-level tendency toward more orthographic variation. 

%This is particularly relevant to our community of practice framework, since we expect the older and more established members of a community to contribute more to the changing norms~\cite{lave1991}.
%\jacob{I still think this expectation is nonintuitive, given that Danescu et al found the opposite, and that in general established individuals are usually thought of as being less favorable towards cultural changes of all kinds. Certainly that's what I would understand from prior work in SLX.}
% We compare this model with one that uses date, the original hashtags, and the relative time variables (``date + originals + relative time'') as predictors. One might expect that a member would employ a variant tag to replace the original banned tag, and thus that the absence of an original tag would be an adequate predictor for a variant tag.

The results of the age regression are displayed in Table~\ref{tab:baselineLogRegressionRelativeTime} and the results of the lifespan regression in Table~\ref{tab:baselineLogRegressionLifespan}.
The date coefficient is consistently positive across regressions, reflecting a community-level trend toward more variants over real time (see \autoref{fig:editDistTime}). 
We also see consistent coefficients for \varname{since\_start} and \varname{till\_end} (negative and positive), showing a coherent member-level trend away from variants over the member's lifespan in the community. 
Taken together, these regressions indicate that orthographic variation is perpetuated by newcomers who bring in the new variants and abandon them over the course of their lifespan. 
The positive coefficient for \varname{date\_range} shows that members who will participate or have participated for longer are more likely to use a variant, suggesting that committed members are more prone to participating in the community change.

\begin{table}[t!]
\begin{center}
%\sffamily
\small
\begin{tabular}{ llll } \toprule
Variable & $\beta$ & $SE$ &  Effect size \\ \midrule
\varname{since\_start} & -0.00456*** & 2.97E-4 & -0.348 \\
\varname{till\_end} & 0.00294*** & 2.88E-4  & 0.654 \\
\varname{date} & 0.00529*** & 1.77E-4 & 0.746 \\ \bottomrule
\end{tabular}
\end{center}
\caption{Regression results for variant appearance in a post, as predicted by relative time variables. 
  *** = $p < 0.0001$. 
In all tables, $\beta$ indicates the regression coefficient and $SE$ indicates the standard error.}
\label{tab:baselineLogRegressionRelativeTime}
\end{table}

\begin{table}[t!]
\begin{center}
\small
%\sffamily
\begin{tabular}{ llll } \toprule
Variable & $\beta$ & $SE$ &  Effect size \\ \midrule
\varname{date\_range} & 0.00294*** & 2.89E-4 & 0.654 \\ 
\varname{date} & 0.00541*** & 1.77E-4 & 0.746 \\ \bottomrule
\end{tabular}
\end{center}
\caption{Regression results for variant appearance in a post, as predicted by the length of a member's lifespan (observed activity period). 
*** = $p < 0.0001$.}
\label{tab:baselineLogRegressionLifespan}
\end{table}

\begin{figure}[t!]
\begin{center}
\includegraphics[width=0.7\columnwidth]{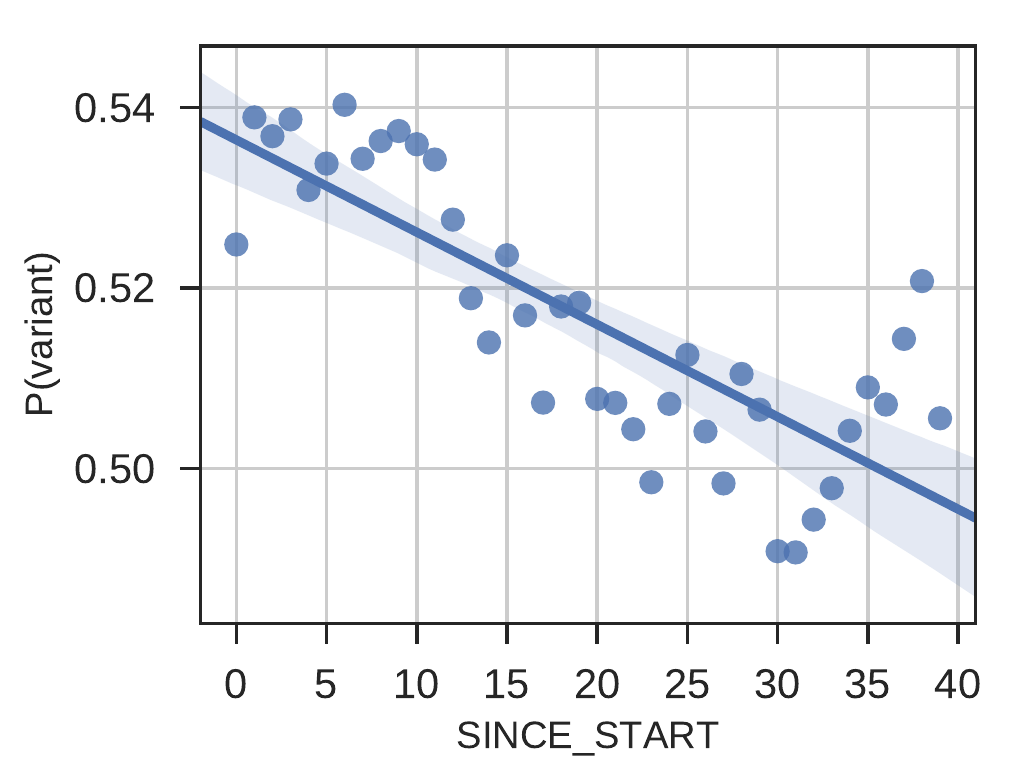}
\caption{Probability of using a variant versus a member's age (weeks since first pro-ED post)
}
\label{fig:variantProbAge}
\end{center}
\end{figure}

\begin{figure*}[t!]
\begin{center}
\includegraphics[width=0.8\textwidth]{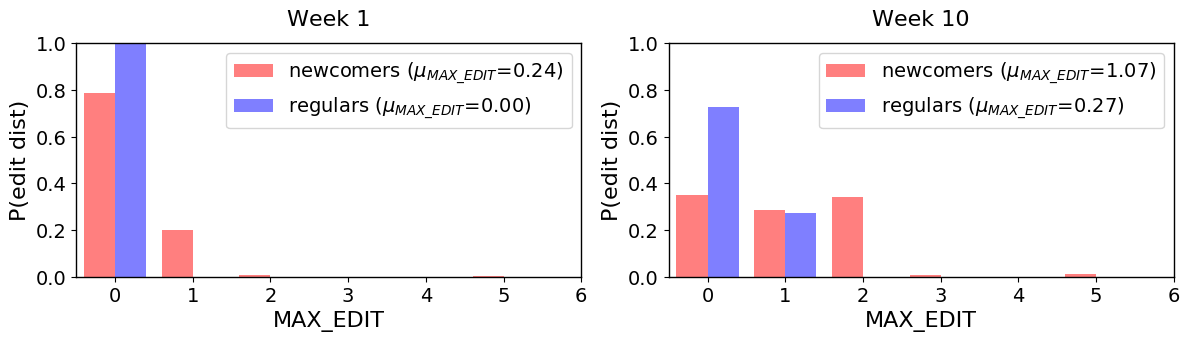}
\end{center}
\caption{Distribution of maximum edit distances across all posts of specified member group (regular versus newcomer) at one week and 10 weeks after the ban (including average edit distance for each group). The newcomers use orthographic variants with consistently higher edit distances than the regulars.}
\label{fig:editDistributionByUser}
\end{figure*}

Both models achieve a better fit than null. 
The deviance of the null model and the deviance of both models approximately follow a $\chi^2$ distribution, with degrees of freedom equal to the number of additional variables in the latter model: for age, $\chi^{2}(3, N=100,000) = 277,258 - 276,280 = 978$, $p<10^{-5}$ and for lifespan, $\chi^{2}(2, N=100,000) = 277,258 - 276,334 = 924$, $p<10^{-5}$.

This analysis uncovers a split between community-level and member-level variant adoption. As time passes, the overall frequency of orthographic variation increases; but as individual members grow older, they are less likely to use variants, as shown in \autoref{fig:variantProbAge}.
%% save this for discussion
%This is consistent with prior work in online language variation~\cite{danescu2013}. 
On the other hand, individuals who post pro-ED content over a long period of time are 4.33\% more likely to use a variant than transient community members ($t=30.9, p < 0.001$).
This difference holds up for the intersection of the two variables: committed newcomers are 5.09\% more likely to use a variant than transient newcomers ($t=25.4, p < 0.001$).
% data_processing/compare_variant_prob_date_range.ipynb
Overall, \emph{committed} members and \emph{newcomers} are the main contributors to the change toward more frequent orthographic variants.

%% We've already reviewed the RQs several times, I don't see the need to do it here. -J
%% I also don't understand introducing social role only in RQ2, when we use the same predictors/IVs in RQ1 -J
%We follow up this finding on frequency of variation with a question on depth of variation: are the deeper orthographic variants more associated with social roles~\cite{bernstein2011}? If so, we would expect to see a stronger effect of the proposed social roles, such that committed newcomers would be especially likely to use deeper orthographic variants. 

\begin{figure}[t!]
\includegraphics[width=0.9\columnwidth]{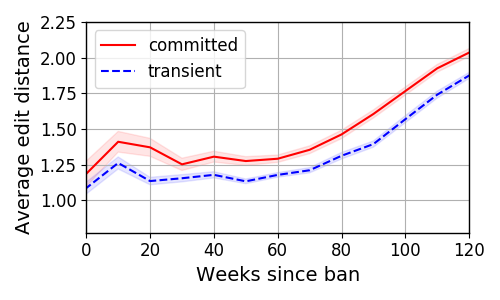}
\caption{Average edit distance over time, binned by \varname{date} and \varname{date\_range} and including 95\% confidence intervals. 
% The transient category contains all members with \varname{date\_range} between 0 and 10 weeks, and the committed category contains all members with \varname{date\_range} greater than 10 weeks. 
%The committed group has a consistently higher average edit distance than the transient group. 
}
\label{fig:binnedEditDistDate}
\end{figure}

\subsection{RQ2: Is depth of variation affected by membership attributes?}
\label{sec:rq2}
%RQ1 considered only the frequency of orthographic variation. 
%By measuring edit distance from the source hashtag, it is possible to ask the same questions regarding depth of variation. 
We now examine whether the social correlates of orthographic variation are stronger for variants that are further from the original spelling. This is done by grouping variants by edit distance, and measuring the strength of association with membership attributes for low and high edit-distance spellings. 
%We first compare these associations directly through univariate analysis, and then perform a set of multivariate regressions.

\subsubsection{Univariate analysis}
\autoref{fig:editDistributionByUser} shows the frequency of posts containing an orthographic variant with edit distances 1-6, broken down by week (since the ban) and by member age (newcomers versus regulars). The newcomers clearly outpace the regular community members in adopting orthographic variants with higher edit distance. Figure~\ref{fig:binnedEditDistDate} demonstrates the impact of member lifespan, comparing the average maximum edit distance in posts from committed and transient members of the pro-ED community. Both transient and committed members follow the same community level trend toward using variants with higher edit distance over time, and the separation between transient and committed members remains robust even two years after the ban. 
%% too small to worry about, let's stay on message -j
%The bump at the beginning of the series around week 10 is likely the result of the popularity spike in deeper variants shortly following the ban date (see Figure~\ref{fig:editDistTime}). 
To confirm the difference between transient and committed members, we tested all split points in the range from 8-12 weeks and found similar results, suggesting that member lifespan can be reliably correlated with orthographic variation. 

% wide table
\begin{table*}[t]
\centering
%\resizebox{\columnwidth}{!}{%
%\begin{minipage}{0.6\textwidth}
%\sffamily
\small
\begin{tabular}{llllllll}
\toprule
Model type & \multicolumn{3}{c}{Dependent variable: \varname{edit\_dist\_1}} & & \multicolumn{3}{c}{\varname{Dependent variable: edit\_dist\_4}} \\
\cmidrule{2-4} \cmidrule{6-8} 
 & $\beta$ & $SE$ & Effect size & \phantom{a} & $\beta$ & $SE$ & Effect size \\ \midrule
\textit{Age} & & & & & & \\ 
\varname{since\_start} & -0.00177*** & 2.98E-4 & -0.097 &  & -0.00450*** & 3.12E-4 & -0.416 \\
\varname{till\_end} & 0.00311*** & 2.85E-4 & 0.250 &  & 0.0133*** & 4.00E-4 & 1.22 \\
\varname{date} & -0.00149*** & 1.75E-4 & -0.127 & & 0.0410*** & 2.84E-4 & 3.85 \\ [4pt]
\textit{Lifespan} & & & & & & \\ 
\varname{date\_range} & 0.00311*** & 2.88E-4 & 0.250 &  & 0.0133*** & 4.03E-4 & 1.22 \\
\varname{date} & -0.00149* & 1.76E-4 & -0.127 &  & 0.0344*** & 2.88E-4 & 3.85 \\ \bottomrule
\end{tabular}%
%\end{minipage}
%}
\caption{Logistic regressions to predict the appearance of a variant with a specified edit distance, as predicted by (1) age and (2) lifespan. 
%We see that the group with a higher edit distance exhibits more extreme correlations with relative time as well as with member lifespan. 
*** = $p < 0.0001$, * = $p < 0.05$.
}
\label{tab:editDistLogRegression}
\end{table*}

\subsubsection{Multiple logistic regression} 
To understand the combined impact of member age and lifespan, we use logistic regression, with the outcome variable indicating whether the post contains an orthographic variant of edit distance 1-4 from the source hashtag.
%Results are shown in Table~\ref{tab:editDistLogRegression}.
%predicting the appearance of variants with the lowest and highest edit distances (1 and 4). 
The results in \autoref{tab:editDistLogRegression} show that effect sizes are larger for the higher edit distance variants, which are more quickly adopted by newcomers, more quickly abandoned by older members, and more strongly favored by committed community members. 
Social differences therefore correlate not only with the frequency of orthographic variation, but also the depth; conversely, these deeper orthographic variables are better indicators of each member's position in the community.

All edit distance regression models achieve a fit significantly better than the null model: e.g., for the edit distance 4 age regression model, the difference of its deviance from that of the null model approximately follows a $\chi^2$ distribution: $\chi^{2}(3, N=2,416,259) = 277,258 - 253,808 = 23,450$, $p<10^{-5}$.

%Outside of the use of orthographic variation by individuals, we must also consider how this variation is received by other Instagram users, in order to determine if such variation is an important signal for group membership \cite{blashki2005}.

\subsection{RQ3: Does orthographic variation affect social reception?}
\label{sec:rq3}
Finally, we investigate how orthographic variation is received by the community using likes and comments received on a post. 
%For each post we have the number of likes and comments at the time of its collection, which represent a post's social reception. 
%by the pro-ED community. 
%% there is no way to ensure that likes and comments are from the pro-ED community; comments could be from anti-ED people, no?
Although Chancellor~\etal~\cite{chancellor2016variation} find that posts with a variant receive more social engagement, it remains to be seen whether this effect is strengthened with deeper edit distance. Since the community norm moves towards variants with deeper edit distance, we expect that posts containing deeper variants would achieve higher engagement in the form of both likes and comments. 

%We use a Poisson regression, predicting the social engagement that a post receives in $likes$ and $comments$. 
To predict the social reception on a given post, we use a Poisson regression, with the outcome variable corresponding to the logarithm of the number of likes and comments for each post.\footnote{We use the \emph{plm} package in R~\cite{croissant2008}.} 
The main predictor is the maximum edit distance of the variants in the post (\varname{max\_edit}).
 In addition, we include a number of control predictors: absolute time (\varname{date}), member age (\varname{since\_start}), presence of hashtag variant in post (\varname{variant}), number of hashtags per post (\varname{tags}), 
%popularity of the most popular hashtag in the post at the time of posting (\varname{max\_pop}), 
and presence of a source hashtag or one of its variants (e.g., a post with \example{\#ana} and a post with \example{\#anaa} each have a 1 for feature \varname{ANA}). 
The hashtag-source variables partly control for post topic, since posts about a more popular topic like anorexia might also garner more social reception.
Lastly, a fixed effect for each member is added to control for the possibility that some members receive more social reception than others, due to higher follower counts.
% \jacob{now would be the time to talk about the fixed effect. what is ``user's social status''? is this one of the ``control predictors'', or something else?}

\input{social-reception-table}

As shown in Table~\ref{tab:socialRegression}, posts with deeper variants (higher \varname{max\_edit}) are positively associated with social engagement through ``likes.''
This complements the earlier finding that posts with variants received more social attention: increased social attention varies with the depth of variation.
%% would need to compare with the counterfactual of how much attention they'd get without the ban
% It also suggests a success for the pro-ED community: despite Instagram's bans, community members were able to generate variants that attracted more social attention. 

However, edit distance is not significantly correlated with comments received, which suggests that posts with especially deep variant hashtags do not elicit the more expensive social signal of a comment (as opposed to the passive ``like'' signal). This may be due to the relatively high proportion of posts with no comments (heavy left-tail of \varname{logcomments} in \autoref{fig:summaryStats}).
As expected, posts with more tags tend to receive more engagement: such posts are easier to find, using Instagram's hashtag-based search functionality. 
Finally, community members tend to gain fewer likes as they ``age'' (positive \varname{since\_start}), possibly because they are actively engaged with other members, or simply because novelty drives interest.

\subsection{Comparison with Twitter}
\label{sec:twitter_comparison}
% compare_twitter_instagram_variation.ipynb#Compare-Twitter-and-Instagram-variation

To better understand how the proliferation of orthographic variation relates to Instagram's restrictions on pro-ED hashtags, we compare with Twitter, which did not implement any restrictions on pro-ED hashtags. 
Twitter is a useful comparison because, like Instagram, members employ hashtags to index their posts by topic, which helps other members search for posts of interest.
We extracted a sample of 4043 tweets containing at least one of the banned or variant hashtags, spanning from January 2012 to June 2014.
%% save for acknowledgments section, and do not include until camera-ready version
% \footnote{We thank Brendan O'Connor for supplying this data.}

\begin{figure}[t!]
\centering
\includegraphics[width=0.8\columnwidth]{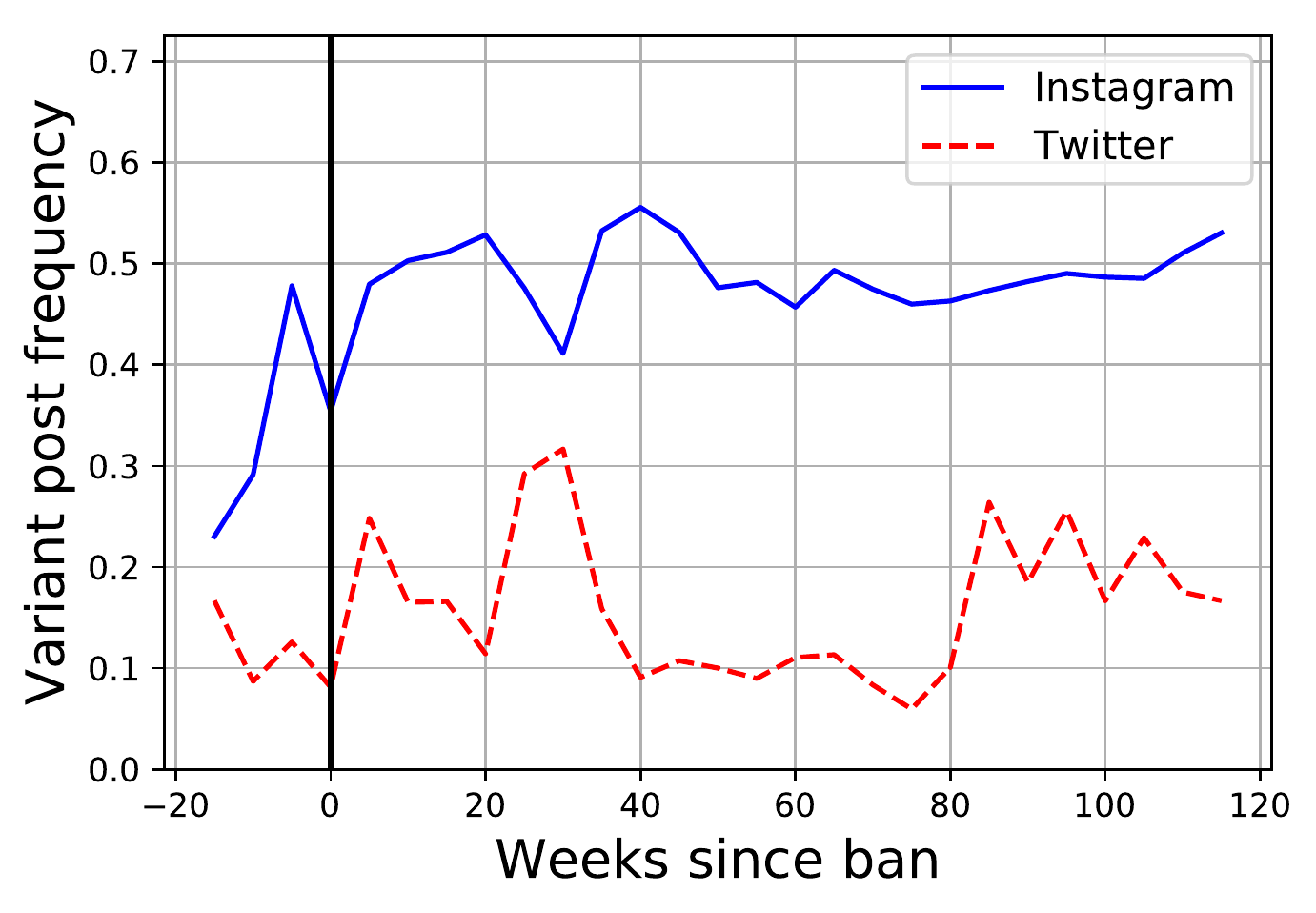}
\caption{Frequency of posts containing at least one hashtag variant, on Instagram and Twitter.}
\label{fig:instagram_twitter_variant_frequency}
\end{figure}

The proportion of variants in this sample is significantly lower than in our Instagram data: 51.9\% of the pro-ED Instagram posts contain at least one variant, while on Twitter only 15.0\% of pro-ED posts contain at least one variant ($Z=46.8, p < 0.001$). 
Furthermore, on Twitter there was no noticeable change in posting behavior after Instagram's actions in April 2012, as shown in \autoref{fig:instagram_twitter_variant_frequency}.
%While the proportion of Instagram posts with at least one variant increases steadily after the ban, the same proportion of Twitter posts remains more or less constant.
The difference in mean variant post frequency before and after the ban is significant in the Instagram data (difference = 13.1\%, $Z=16.4, p < 0.001$) but not significant in the Twitter data (difference = 3.4\% $Z=-1.64, p > 0.05$).
%In summary, we find that there was likely no change in hashtag variation on Twitter corresponding to the content ban on Instagram.
The orthographic variation on Twitter was also considerably less diverse: of the 608 tweets containing a variant hashtag, only 23 tweets contained a variant hashtag with edit distance greater than one.
%\ian{could include a bar chart comparing the distribution of variants over edit distance, for Twitter and Instagram}
None of those 23 tweets reached Twitter's 140 character limit (mean character count 70.3), suggesting that the lack of orthographic variation on Twitter was not due to the character limit.
This cross-community comparison demonstrates that the pro-ED hashtags were more likely a result of the content ban on Instagram than a result of an overall trend toward more variation across social media platforms.
%% difference in demographics doesn't explain the temporal change in instagram, unless demographics changed at the same time! -J
% (although there may be a difference in user demographics; see \autoref{sec:limitations_future}).
%However, we acknowledge that we cannot control for points of difference between Instagram and Twitter, such as user demographics
%, which prevents us from comparing the role of orthographic variation as a community marker.
%\vspace{-10pt}

%% file: social-reception-table.tex
\begin{table}[t!]
\begin{center}
\begin{tabular}{ lll} \toprule
 & $\beta$ & $SE$ \\ 
\midrule
\multicolumn{3}{l}{\textit{Dependent variable}: \varname{Logcomments}}  \\ 
\varname{since\_start} & 5.27E-3* & 1.57E-3 \\ 
\varname{tags} & 0.110*** & 2.57E-3 \\
\varname{variant} & -7.89E-3 & 3.44E-3 \\ 
\varname{max\_pop} & -2.33E-3 & 1.26E-3 \\
\varname{max\_edit} & -3.716E-3 & 5.51E-3 \\ [8pt] %\bottomrule
\multicolumn{3}{l}{\textit{Dependent variable}: \varname{Loglikes}}  \\ 
\varname{since\_start} & -0.0319*** & 9.03E-4 \\ 
\varname{tags} & 0.224*** & 1.47E-3 \\
\varname{variant} & -1.14E-3 & 1.98E-3 \\ 
\varname{max\_pop} & -3.89E-3*** & 7.25E-4 \\
\varname{max\_edit} & 0.0130*** & 3.16E-3 \\ \bottomrule
\end{tabular}
\end{center}
\caption{Poisson regressions for social reception, as predicted by membership and language variables (hashtag coefficients omitted for brevity). 
*** = $p < 0.0001$, otherwise $p > 0.05$. 
Both models achieve a weak fit: the \varname{logcomments} regression has $R^{2}$=6.82E-3 ($F=107, p < 0.001$) and the \varname{loglikes} has $R^{2}$=0.0902 ($F=1550, p < 0.001$).
%Both models achieve a weak fit: when compared to the null (intercept-only) model, the \varname{logcomments} model has a $\chi^{2}$ effect size of 0.195 and the \varname{loglikes} model has an effect size of 0.125. 
}
\label{tab:socialRegression}
\end{table}

%% file: discussion.tex
\section{Discussion}

\subsection{Theoretical and Practical Implications}
\label{sec:implications}
Our main finding --- that committed newcomers led the change towards increased orthographic variation --- shows that changing community practices can be tied to the members' progression from newcomers to regulars. 
It is important to relate this result to the earlier finding that newcomers adopt innovative language practices, and then retain these practices even as they become outdated with respect to the rest of the community~\cite{danescu2013,tagliamonte2007}.
The ``old-timers'' in the beer forums studied by Danescu-Niculescu-Mizil \etal~\cite{danescu2013} are merely conservative, clinging to the linguistic habits of their youth. 
In contrast, pro-ED Instagram members were regressive: they began with innovative practices, but they abandoned these practices and returned to standard spellings --- even as the overall community change was driven by subsequent waves of newcomers toward ever more frequent and deeper orthographic variation.

Our work uses orthographic variation as a lens to measure an individual's linguistic distance from an established standard language, a distance which reflects how members of a community can distinguish themselves from others~\cite{eckert1992}. 
This adds another method to the toolkit of language analysis and provides an interesting path for future work in social computing: can the behavior of members of a community be characterized along a continuum, by their linguistic distance from standard language? 
For example, in a community with relatively standard writing practices, the use of excessive capitalization and lengthening (e.g., \example{duuuuudddde}) may be viewed as a non-conformist position towards the community, in comparison with more mild examples of expressive lengthening (e.g., \example{duude}). The opposite can also apply: in a freewheeling community like 4chan, purposeful misspellings may be more common and the use of conventional orthography might be viewed as deviation from the community standard~\cite{bernstein2011}.

\subsection{Limitations and future work}
\label{sec:limitations_future}
%Our study is limited by several factors that point toward future research directions. 
Because Instagram's content ban prevented us from collecting the data directly (e.g. querying for banned terms), we may have missed some orthographic variants. 
Furthermore, Instagram's API prevented us from querying for additional member information, such as the date at which each member joined the site instead of the first date at which they used a pro-ED hashtag. 
This information would complement our analysis and allow us to differentiate newcomers from regulars based on their actual first post date.
Having more detailed member information would also provide a better perspective on the correlation between orthographic variation and social reception: for example, we would be able to test for a connection between social network structure and orthographic variation.
%% sorry, can't understand this sentence -J
% This would help to control for the user-specific effects of social reception, for which we would be unable to account without more rigorous matching techniques.
With respect to the cross-platform comparison, there are important differences in demographics between platforms: as of 2014 Instagram skewed slightly more toward women than Twitter~\cite{duggan2015}, which could result in different aggregate behaviors between the platforms. 
% relevant because eating disorders also disproportionately affect women in the U.S.~\cite{wade2011}.
%However, more rich user information will not help to detect cases in which individuals delete or abandon their accounts, and then return to the community with a new account.

Future work may explore three possible explanations for the role of newcomers.
First, the new pro-ED community community members may adopt the most extreme practices to signal legitimacy in the community, which represents an extreme version of the Community of Practice model in which members gain legitimacy through adoption of social and linguistic practices~\cite{lave1991}.
Second, the adoption of more extreme hashtag variants may represent a form of ``flag-planting,'' by which a newcomer attempts to claim a particular hashtag as their own with an especially extreme variant.
Third, the supposed ``newcomer'' members could actually be new accounts created as a result of being banned, who then adopt more extreme variants to avoid being banned again.
This third possibility is especially relevant in the face of prior findings that moderation of deviant behavior online may cause the deviant community member's practices to become more extreme~\cite{cheng2015}.

%% file: conclusion.tex
\section{Conclusion}
Our study uses orthographic variation to characterize community-level change and differentiating community members by social role. 
A community-level change toward more orthographic variation is driven by committed newcomers, who later abandon their use of variants and accordingly receive more social response.
Furthermore, the depth of orthographic variation differentiates members by age and lifespan, and can weakly predict the level of social response that a post receives. 
%Lastly, by comparing the findings with Twitter data we see that the pro-ED orthographic variation was more likely related to Instagram's content ban than to a general trend toward more variation.
These results have the potential to push social computing research to consider a wider range of language variation, outside of typical change such as adoption of slang, when characterizing an online community and the behavior of its members. 
Employing non-lexical metrics like orthographic edit distance can help researchers capture linguistic change that may otherwise not be apparent.

%% file: acknowledgments.tex
\section*{Acknowledgments}
We thank the anonymous reviewers for their feedback, and the audience at the Diversity and Variation in Language Conference at Emory for their feedback on an early version of this work.
We also thank Brendan O'Connor for providing the Twitter data for comparison in \autoref{sec:twitter_comparison}.  This research was supported by Air Force Office of Scientific Research award FA9550-14-1-0379, by National Institutes of Health award R01-GM112697, and by the National Science Foundation award 1452443.

%% file: main.bbl
\begin{thebibliography}{10}
\providecommand{\url}[1]{#1}
\csname url@samestyle\endcsname
\providecommand{\newblock}{\relax}
\providecommand{\bibinfo}[2]{#2}
\providecommand{\BIBentrySTDinterwordspacing}{\spaceskip=0pt\relax}
\providecommand{\BIBentryALTinterwordstretchfactor}{4}
\providecommand{\BIBentryALTinterwordspacing}{\spaceskip=\fontdimen2\font plus
\BIBentryALTinterwordstretchfactor\fontdimen3\font minus
  \fontdimen4\font\relax}
\providecommand{\BIBforeignlanguage}[2]{{%
\expandafter\ifx\csname l@#1\endcsname\relax
\typeout{** WARNING: IEEEtran.bst: No hyphenation pattern has been}%
\typeout{** loaded for the language `#1'. Using the pattern for}%
\typeout{** the default language instead.}%
\else
\language=\csname l@#1\endcsname
\fi
#2}}
\providecommand{\BIBdecl}{\relax}
\BIBdecl

\bibitem{bernstein2011}
M.~Bernstein, A.~Monroy-Hern{\'{a}}ndez, D.~Harry, P.~Andr{\'{e}}, K.~Panovich,
  and G.~Vargas, ``{4chan and /b/: An Analysis of Anonymity and Ephemerality in
  a Large Online Community},'' \emph{Proceedings of ICWSM}, pp. 50--57, 2011.

\bibitem{bryant2005}
S.~L. Bryant, A.~Forte, and A.~Bruckman, ``{Becoming Wikipedian: transformation
  of participation in a collaborative online encyclopedia},'' in
  \emph{Proceedings of GROUP}, 2005, pp. 1--10.

\bibitem{lave1991}
J.~Lave and E.~Wenger, \emph{Situated learning: Legitimate peripheral
  participation}.\hskip 1em plus 0.5em minus 0.4em\relax Cambridge University
  Press, 1991.

\bibitem{labov2001}
W.~Labov, \emph{Principles of Linguistic Change: Social Factors}.\hskip 1em
  plus 0.5em minus 0.4em\relax Wiley-Blackwell, 2001, vol.~2.

\bibitem{androutsopoulos2011}
J.~Androutsopoulos, ``Language change and digital media: a review of
  conceptions and evidence,'' in \emph{Standard Languages and Language
  Standards in a Changing Europe}.\hskip 1em plus 0.5em minus 0.4em\relax Novus
  Press, 2011, pp. 145--159.

\bibitem{danescu2013}
C.~Danescu-Niculescu-Mizil, R.~West, D.~Jurafsky, J.~Leskovec, and C.~Potts,
  ``No country for old members: User lifecycle and linguistic change in online
  communities,'' in \emph{{Proceedings of WWW}}, 2013, pp. 307--318.

\bibitem{kooti2012emergence}
F.~Kooti, H.~Yang, M.~Cha, P.~K. Gummadi, and W.~A. Mason, ``The emergence of
  conventions in online social networks.'' in \emph{Proceedings of ICWSM},
  2012.

\bibitem{eisenstein2014}
J.~Eisenstein, B.~O'Connor, N.~A. Smith, and E.~P. Xing, ``{Diffusion of
  lexical change in social media},'' \emph{PLoS ONE}, vol.~9, no.~11, 2014.

\bibitem{chancellor2016variation}
S.~Chancellor, J.~A. Pater, T.~Clear, E.~Gilbert, and M.~De~Choudhury,
  ``\#thyghgapp: Instagram content moderation and lexical variation in
  pro-eating disorder communities,'' in \emph{{Proceedings of CSCW}}, 2016, pp.
  1201--1213.

\bibitem{hiruncharoenvate2015}
C.~Hiruncharoenvate, Z.~Lin, and E.~Gilbert, ``{Algorithmically Bypassing
  Censorship on Sina Weibo with Nondeterministic Homophone Substitutions},'' in
  \emph{Proceedings of ICWSM}, 2015, pp. 150--158.

\bibitem{hasan2012}
H.~Hasan, ``Instagram bans thinspo content,'' \emph{Time Newsfeed}, April 2012.

\bibitem{ren2011}
Y.~Ren, R.~Kraut, S.~Kiesler, and P.~Resnick, ``{Encouraging commitment in
  online communities},'' in \emph{{Evidence-based social design: Mining the
  social sciences to build online communities}}.\hskip 1em plus 0.5em minus
  0.4em\relax MIT Press, 2011, pp. 77--125.

\bibitem{dube2006}
L.~Dub{\'e}, A.~Bourhis, and R.~Jacob, ``Towards a typology of virtual
  communities of practice,'' \emph{Interdisciplinary Journal of Information,
  Knowledge, and Management}, vol.~1, no.~1, pp. 69--93, 2006.

\bibitem{schlager2002}
M.~Schlager, J.~Fusco, and P.~Schank, ``Evolution of an online education
  community of practice,'' in \emph{Building virtual communities: Learning and
  change in cyberspace}.\hskip 1em plus 0.5em minus 0.4em\relax Cambridge
  University Press, 2002, pp. 129--158.

\bibitem{blashki2005}
K.~Blashki and S.~Nichol, ``Game geek's goss: linguistic creativity in young
  males within an online university forum,'' \emph{Australian Journal of
  Emerging Technologies and Society}, vol.~3, no.~2, pp. 71--80, 2005.

\bibitem{cobb2016}
G.~Cobb, ``{``This is not pro-ana'': Denial and disguise in pro-anorexia online
  spaces},'' \emph{Fat Studies}, vol.~5, no.~2, pp. 1--17, 2016.

\bibitem{manikonda2017}
L.~Manikonda and M.~De~Choudhury, ``Modeling and characterizing visual
  attributes of mental health disclosures in social media,'' \emph{Proceedings
  of CHI}, 2017.

\bibitem{eckert1992}
P.~Eckert and S.~McConnell-Ginet, ``{Think Practically and Look Locally:
  Language and Gender as Community-Based Practice},'' \emph{Annual Review of
  Anthropology}, vol.~21, pp. 461--490, 1992.

\bibitem{labov2007}
W.~Labov, ``Transmission and diffusion,'' \emph{Language}, vol.~83, no.~2, pp.
  344--387, 2007.

\bibitem{eisenstein2013}
J.~Eisenstein, ``What to do about bad language on the internet,'' in
  \emph{Proceedings of HLT-NAACL}, 2013, pp. 359--369.

\bibitem{herring2012}
S.~C. Herring, ``{Grammar and electronic communication},'' \emph{The
  Encyclopedia of Applied Linguistics}, pp. 1--9, 2012.

\bibitem{jaffe2012orthography}
A.~Jaffe, J.~Androutsopoulos, M.~Sebba, and S.~Johnson, \emph{Orthography as
  Social Action: Scripts, Spelling, Identity and Power}.\hskip 1em plus 0.5em
  minus 0.4em\relax Berlin: Walter de Gruyter, 2012.

\bibitem{eisenstein2015systematic}
J.~Eisenstein, ``Systematic patterning in phonologically-motivated orthographic
  variation,'' \emph{{Journal of Sociolinguistics}}, vol.~19, pp. 161--188,
  2015.

\bibitem{tsur2015}
O.~Tsur and A.~Rappoport, ``Don't let me be \#misunderstood: Linguistically
  motivated algorithm for predicting the popularity of textual memes,'' in
  \emph{Proceedings of ICWSM}, 2015, pp. 426--435.

\bibitem{zelenkauskaite2008}
A.~Zelenkauskaite and S.~C. Herring, ``{Television-mediated conversation:
  Coherence in Italian iTV SMS chat},'' in \emph{{Proceedings of HICSS}}, 2008,
  pp. 145--145.

\bibitem{schnoebelen2012}
T.~Schnoebelen, ``{Do you smile with your nose? Stylistic variation in Twitter
  emoticons},'' \emph{University of Pennsylvania Working Papers in
  Linguistics}, vol.~18, no.~2, p.~14, 2012.

\bibitem{back2013}
M.~Back and M.~Zepeda, ``{Performing and positioning orthography in Peruvian
  CMC},'' \emph{Journal of Computer-Mediated Communication}, vol.~18, no.~2,
  pp. 119--135, 2013.

\bibitem{leigh2009}
C.~Leigh, ``Lurkers and lolcats: An easy way from out to in,'' \emph{{Journal
  of Digital Research and Publishing}}, vol.~2, pp. 131--139, 2009.

\bibitem{levenshtein1966binary}
V.~Levenshtein, ``Binary codes capable of correcting deletions, insertions and
  reversals,'' in \emph{{Soviet Physics Doklady}}, vol.~10, 1966, pp. 707--710.

\bibitem{heeringa2006}
W.~Heeringa, P.~Kleiweg, C.~Gooskens, and J.~Nerbonne, ``Evaluation of string
  distance algorithms for dialectology,'' in \emph{{Proceedings of ACL}}, 2006,
  pp. 51--62.

\bibitem{gardner1995regression}
W.~Gardner, E.~P. Mulvey, and E.~C. Shaw, ``{Regression analyses of counts and
  rates: Poisson, overdispersed Poisson, and negative binomial models},''
  \emph{Psychological bulletin}, vol. 118, no.~3, pp. 392--404, 1995.

\bibitem{chinn2000}
S.~Chinn, ``A simple method for converting an odds ratio to effect size for use
  in meta-analysis,'' \emph{Statistics in medicine}, vol.~19, no.~22, pp.
  3127--3131, 2000.

\bibitem{croissant2008}
Y.~Croissant, G.~Millo \emph{et~al.}, ``Panel data econometrics in r: The plm
  package,'' \emph{Journal of Statistical Software}, vol.~27, no.~2, pp. 1--43,
  2008.

\bibitem{tagliamonte2007}
S.~A. Tagliamonte and A.~D'Arcy, ``{Frequency and variation in the community
  grammar: Tracking a new change through the generations},'' \emph{Language
  Variation and Change}, vol.~19, no.~02, pp. 199--217, 2007.

\bibitem{duggan2015}
M.~Duggan, N.~Ellison, C.~Lampe, A.~Lenhart, and M.~Madden, ``Social media
  update 2014. pew research center, 2015,'' 2015.

\bibitem{cheng2015}
J.~Cheng, C.~Danescu-Niculescu-Mizil, and J.~Leskovec, ``{Antisocial Behavior
  in Online Discussion Communities},'' in \emph{Proceedings of ICWSM}, 2015.

\end{thebibliography}
